\documentclass[10pt,twocolumn,letterpaper]{article}

\usepackage[pagenumbers]{cvpr} 

\usepackage{graphicx}
\usepackage{amsmath}
\usepackage{amssymb}
\usepackage{booktabs}
\usepackage{multirow}
\usepackage{algorithm}
\usepackage{algorithmic}
\usepackage[hyphens]{url} 
\usepackage{dsfont}
\usepackage[pagebackref,breaklinks,colorlinks]{hyperref}

\usepackage[capitalize]{cleveref}
\crefname{section}{Sec.}{Secs.}
\Crefname{section}{Section}{Sections}
\Crefname{table}{Table}{Tables}
\crefname{table}{Tab.}{Tabs.}

\def\cvprPaperID{1640} 
\def\confName{CVPR}
\def\confYear{2022}

\begin{document}

\title{ImpDet: Exploring Implicit Fields for 3D Object Detection}

\author{Xuelin Qian$^{1}$ \quad Li Wang$^{1}$ \quad Yi Zhu$^{2,}$\thanks{Work done outside Amazon} \quad Li Zhang$^{1}$ \quad Yanwei Fu$^{1}$ \quad Xiangyang Xue$^{1}$ \\
$^{1}$Fudan University \quad $^{2}$Amazon Inc.}

\maketitle

\begin{abstract}

Conventional 3D object detection approaches concentrate on bounding boxes representation learning with several parameters, i.e., localization, dimension, and orientation. Despite its popularity and universality, such a straightforward paradigm is sensitive to slight numerical deviations, especially in localization. By exploiting the property that point clouds are naturally captured on the surface of objects along with accurate location and intensity information, we introduce a new perspective that views bounding box regression as an implicit function. This leads to our proposed framework, termed Implicit Detection or ImpDet, which leverages implicit field learning for 3D object detection. Our ImpDet assigns specific values to points in different local 3D spaces, thereby high-quality boundaries can be generated by classifying points inside or outside the boundary. To solve the problem of sparsity on the object surface, we further present a simple yet efficient virtual sampling strategy to not only fill the empty region, but also learn rich semantic features to help refine the boundaries. Extensive experimental results on KITTI and Waymo benchmarks demonstrate the effectiveness and robustness of unifying implicit fields into object detection.

\end{abstract}

\section{Introduction}
\label{sec:intro}
3D object detection has attracted substantial attention in both academia and industry due to its wide applications in autonomous driving~\cite{DBLP:conf/cvpr/GeigerLU12,sun2020scalability,DBLP:conf/cvpr/CaesarBLVLXKPBB20}, virtual reality~\cite{park2008multiple,mathis2021fast} and robotics~\cite{chan2021lidar}. Although point clouds generated from 3D LiDAR sensors capture precise distance measurements and geometric information of surrounding environments, the irregular, sparse and orderless properties make it hard to be encoded and non-trivial to directly apply 2D detection methods~\cite{tian2019fcos}. 

\begin{figure}[tbp]
\begin{centering}
\includegraphics[scale=0.38]{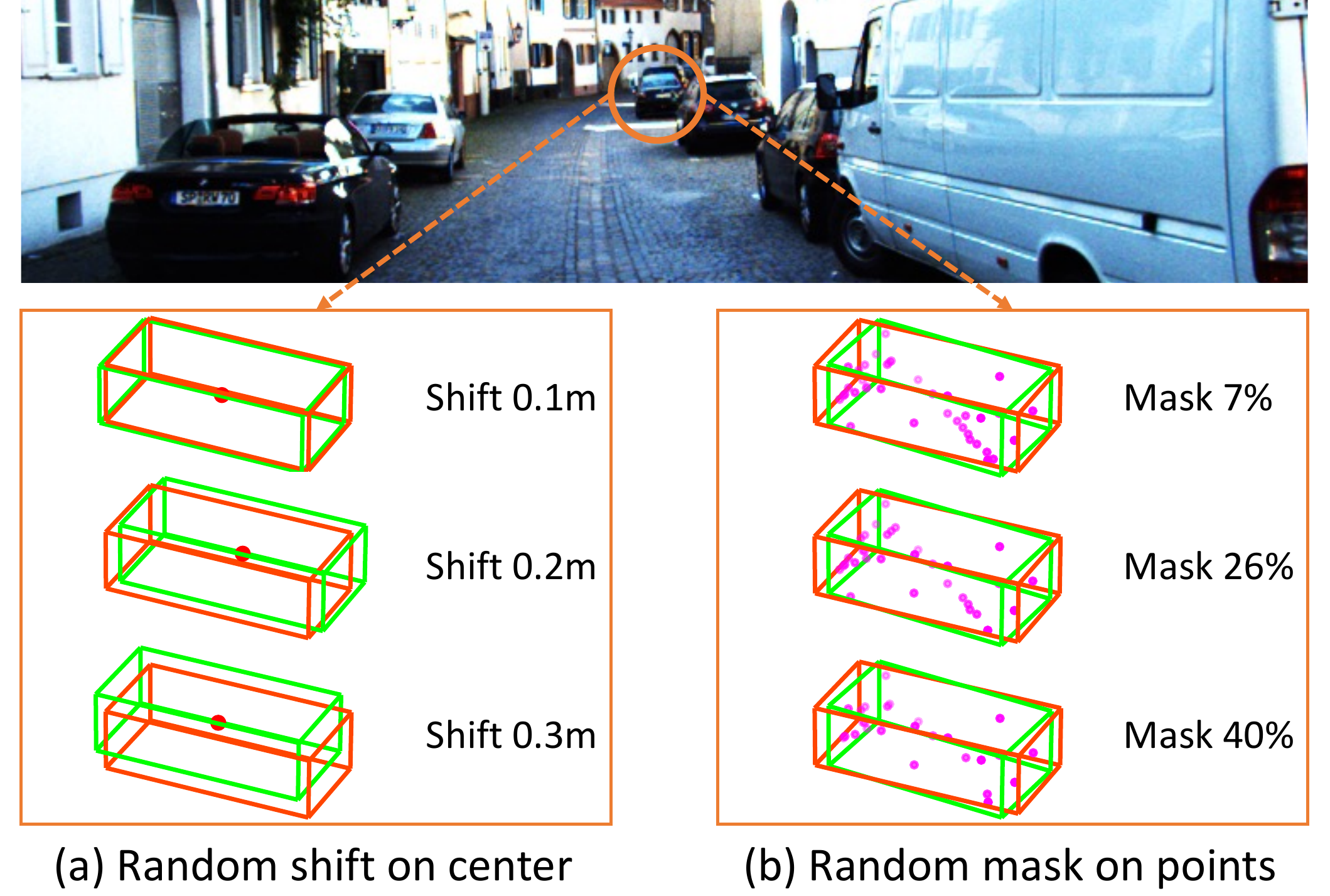}
\caption{
Illustration of different 3D bounding box representations under numerical deviations.
Ground truth and deviated boxes are drawn in red and green respectively. \textbf{(a) Parameters:} random shift ground-truth centers in range $\pm\left(0.1, 0.2, 0.3\right)m$
along x/y/z axis. 
\textbf{(b) Implicit fields:} random mask $7$/$19$/$40\%$ predicted inside points.
We show boxes represented with implicit fields are more robust than conventional parameters when facing some outliers.
}
\label{fig:intro}
\vspace{-0.2in}
\end{centering}
\end{figure}

Generally, object bounding boxes in 3D scenes are represented with several parameters, such as center localization, box dimension, and orientation. Previous literatures~\cite{shi2020pv,yan2018second,yoo20203d,li2021sienet,zheng2021se,li2021p2v} are mostly built upon this representation and utilize convolutional neural networks (CNN) to regress these values. Nevertheless, 
when there are fewer points on objects caused by object occlusion or other factors for sparsity, directly learning  these parameters would be fragile. Even worse, several studies~\cite{ma2021delving,wang2021progressive} have demonstrated that even minor numerical deviations of these parameters may cause significant performance drop, as shown in Fig.~\ref{fig:intro} (a).
Consequently, this motivates us to consider an open question:
\textit{Can we have the more robust 3D bounding box representations for learning?}

Interestingly, recent learning based 3D object modeling works~\cite{chen2019learning,mescheder2019occupancy} employ as the nature recipe the implicit fields, which nevertheless has less touched in 3D object detection.
Thus to nicely answer the above question, this paper particularly highlights the potential of exploiting implicit fields for  3D object detection.
More precisely, implicit field assigns a value (\textit{e.g.}, $0$ or $1$) to each point in the 3D space; then the object's mesh can be represented by all points assigned to a specific value. Inspired by this, we advocate an implicit way to build bounding boxes for object detection, since point clouds are naturally captured on the surface of objects, with accurate location and intensity information. More precisely, we first classify/assign points into two categories, \textit{i.e.}, inside or outside the box. Then, we can fit a bounding box directly according to these points. As illustrated in Fig.~\ref{fig:intro} (b), compared with the conventional box representation, such an implicit way can benefit from the best of both worlds: (1) providing high-quality boxes without any pre-defined anchor and being more robust even to some outliers; (2) naturally leveraging implicit fields for multi-task learning, improving features with point-based representation; (3) effectively enhancing the features of inside points and suppressing the outside points according to the implicit assignments.

\begin{figure*}[thb]
\begin{centering}
\includegraphics[width=0.97\linewidth]{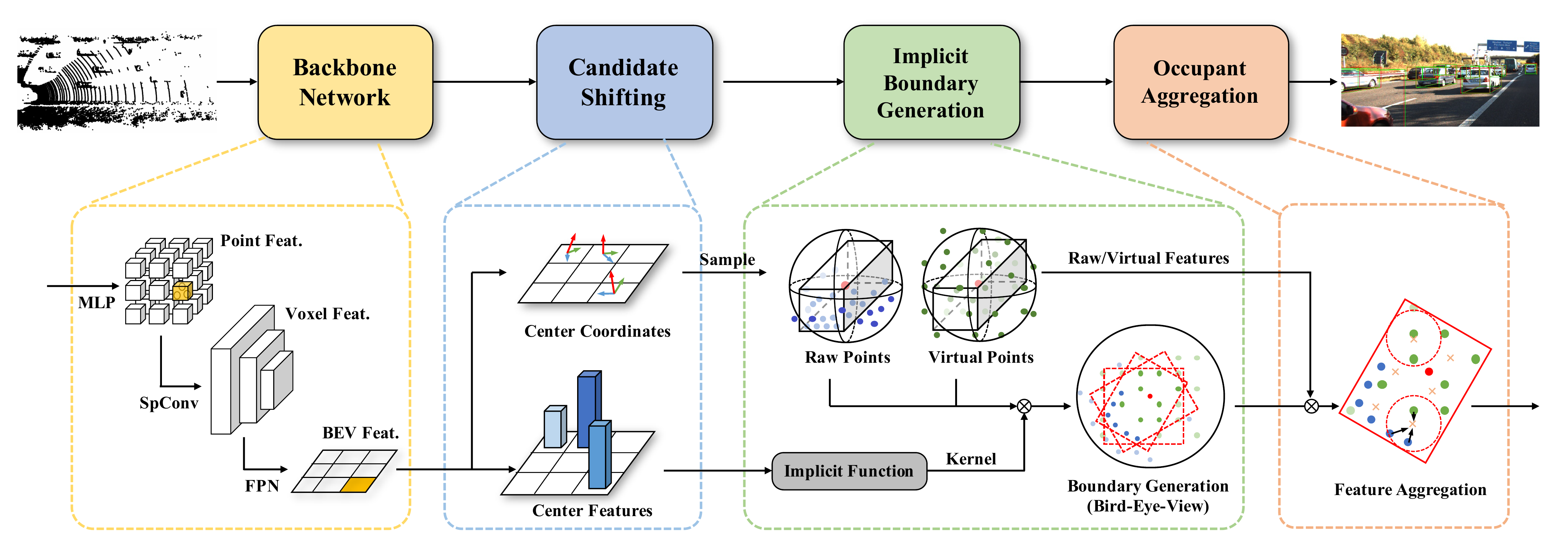}
\caption{An overview of our proposed ImpDet. After obtaining point- and voxel-wise features from the backbone network, the \emph{candidate shifting} module first samples points closest to the ground-truth centers as candidates. Next, we propose an implicit function to fit high-quality boundary boxes by assigning implicit values to classify inside and outside points. A virtual sampling strategy is further introduced to not only fill the empty region in objects but also learn rich semantic features. Finally, we perform the \emph{occupant aggregation} to refine the boundaries by aggregating the feature of points that occupy the inside of boundaries.}
\label{fig:framework}
\end{centering}
\vspace{-0.1in}
\end{figure*}

This paper, for the first time, systematically explores the implicit field learning for 3D object detection, and proposes the ImpDet. As shown in Fig.~\ref{fig:framework}, our ImpDet mainly consists of three key components: (1) candidate shifting, (2) implicit boundary generation and (3) occupant aggregation. 
Specifically, the \emph{candidate shifting} first samples points closest to the ground-truth centers as candidates, in order to relieve the computational pressure caused by implicit functions. Different from previous 3D object detectors explicitly regressing box parameters based on candidates, \emph{implicit boundary generation} allocates a local 3D space for each candidate and then adopts the implicit function to fit high-quality boundaries by assigning implicit values to classify inside and outside points. Furthermore, we come up with a refinement strategy, termed \emph{occupant aggregation}, to refine the boundaries by aggregating features of inside points. Finally, we output the parameter-based representation for detection evaluation.

In summary, our primary contributions are listed as:
(1) We for the first time show a perspective of incorporating implicit fields into 3D object detection and propose a framework named ImpDet. Different from previous detectors explicitly regressing box parameters, our ImpDet uses the implicit function to assign values to each point and then fit high-quality boundaries without any pre-defined anchor.
(2) We propose a simple yet effective virtual sampling strategy to assist the implicit boundary generation since points in objects may be incompleted due to occlusion or sparsity. With multi-task learning, it can not only fill the empty region, but also learn rich semantic information as auxiliary features.
(3) Extensive experiments are conducted on KITTI and Waymo benchmarks to demonstrate the effectiveness and robustness of our ImpDet.

\section{Related Work}
\label{sec:relate}


\noindent \textbf{3D Mesh Representation.} 
There are two commonly used implicit functions for 3D mesh representations, signed distance functions (SDF) and occupancy functions.
For SDF, values inside the shape are negative, and then increase to zero as points approach the boundary, and become positive when points are outside the shape.
Occupancy functions classify points into two categories, $0$ for being inside and $1$ for being outside.
Previous studies~\cite{mescheder2019occupancy,park2019deepsdf,jiang2020local,chibane2020implicit,ibing20213d} have been proposed to utilize CNNs to predict a value for each point of the 3D scene. 
Then, methods like Marching Cubes~\cite{lorensen1987marching} can be used to extract a surface based on both functions.
Given the simplicity of binarized representation, we adopt the occupancy functions as an implicit way to build bounding boxes for 3D object detection. Compared to the conventional box representation, our method provides high-quality boxes without any pre-defined anchor and is more robust even with some outliers.

\vspace{0.05in}
\noindent \textbf{3D Object Detection.}
Although image-based object detection has achieved remarkable progress, it is far from meeting the requirements for real-world applications, such as autonomous driving. 
Therefore, researches on 3D data are gradually emerging and flourishing. Most existing 3D object detection methods can be classified in two directions, \textit{i.e.}, point-based and voxel-based.
Point-based methods~\cite{qi2017pointnet,shi2019pointrcnn,Yang20203dssd,yang2019std} take raw point clouds as input and extract local features with set abstraction. However, the sampling and grouping operations in set abstraction make it time-consuming. 
For voxel-based approaches \cite{shi2020pv,du2020associate,deng2020voxel,zheng2021se,he2020structure}, they divide point clouds into regular grids so that 3D CNNs can be applied for feature extraction. In this work, we adopt the voxel-based CNN as the backbone in consideration of its efficiency.

\vspace{0.05in}
\noindent \textbf{3D Object Detection with Segmentation Branch.}
As another important branch for 3D scene understanding, instance segmentation is gradually applied to assist 3D object detection on account of no cost for annotation. 
\cite{he2020structure,zhong2021vin} adds another segmentation branch as an auxiliary network to guide the features to be aware of object structures. \cite{zhou2020joint,shi2020pv,xie2020pi,shi2019pointrcnn} propose to utilize segmentation results to re-weight features or vote the predicted boxes for refinement. \cite{wang2021pointaugmenting,xie2020pi,vora2020pointpainting,chen2020panonet3d} obtain segmentation labels/features from 2D space to enhance the point representations in the 3D space. 
Methods on this line mostly use simple fully-connected layers to build the extra segmentation branch, except that \cite{zhong2021vin} introduces the concept of implicit function.
Different from existing works, 
we propose a novel unified 3D object detection framework, which for the first time directly benefits from the implicit field learning to achieve more precise 3D object detection. Such a framework attempts to assign a special value for each point via implicit functions. Then the network is able to make full use of the assignment results to provide high-quality boundaries and leverage more discriminative inside features (natural by-product) for refinement.


\section{Methodology}

In this section, we introduce our implicit 3D object detection framework (ImpDet), as illustrated in Fig.~\ref{fig:framework}.
We first describe the backbone network for feature extraction in Sec.~\ref{subsec:backbone}. Then we elaborate three proposed modules for implicit object detection, \textit{candidate shifting}, \textit{implicit boundary generation}, and \textit{occupant aggregation} (Sec.~\ref{subsec:candidate} $\sim$ \ref{subsec:aggregation}).
At last, we discuss the loss functions for model training in Sec.~\ref{subsec:loss}.

\subsection{Backbone Network}
\label{subsec:backbone}

We adopt the voxel-based CNN as the backbone due to its efficiency. In order to prevent the loss of geometry information, which is crucial for implicit boundary generation, we simultaneously extract point-wise and voxel-wise features in one backbone \cite{miao2021pvgnet,zhou2018voxelnet}. As the yellow block shown in Fig.~\ref{fig:framework}, we first feed raw point clouds $\mathcal{P} = \left\{x_{i}, y_{i}, z_{i}, r_{i}\right\}_{i=1}^{N}$ into a MLP for point embedding, where $\left(x_{i}, y_{i}, z_{i} \right)$ and $r_{i}$ mean the coordinates and intensity of point $p_{i}$, $N$ is the total number of points. Then, we utilize stacked VFE layers \cite{zhou2018voxelnet} to obtain the initial features $f^{\left(v_{0}\right)}$ of each voxel. 
The point-wise features $f^{\left(point\right)} \in \mathbb{R}^{N \times 64}$ are subsequently calculated by another MLP layer with the input of point embedding features and $f^{\left(v_{0}\right)}$.
For voxel-wise features, $f^{\left(v_{0}\right)}$ is followed by several 3D sparse convolution blocks to gradually produce multi-scale features $f^{\left(v_{i}\right)} \arrowvert_{i=1}^{5}$. Similar to \cite{deng2020voxel}, we compress the voxel-wise tensor $f^{\left(v_{5}\right)}$ by concatenating features along $z$-axis, and further apply a FPN \cite{lin2017feature} structure to get 2D Bird-Eye-View map features $f^{\left(bev\right)}\in \mathbb{R}^{H \times W \times C}$, where $H$ and $W$ represent the length and width respectively.

\subsection{Candidate Shifting}
\label{subsec:candidate}

This stage first shifts points on BEV maps toward the centers of their corresponding ground-truth boxes and then sample those closest to the centers. The goal is to dramatically reduce the computational costs for the following stages by shifting and sampling points, which is different from \cite{Yang20203dssd}.

Concretely, we use a MLP layer to generate the central offset $p^{\left(ofs\right)} \in \mathbb{R}^{HW \times 3}$ as well as the feature offset $f^{\left(ofs\right)} \in \mathbb{R}^{HW \times C}$ of each pixel on BEV maps. By adding offsets, the candidate centers can be generated as,

\begin{equation}
\left[p^{\left(ofs\right)};~f^{\left(ofs\right)}\right] = \mathcal{M}\left( f^{\left(bev\right)} \right)
\end{equation}
\vspace{-0.3in}

\begin{equation}
p^{\left(ctr\right)} = p^{\left(ofs\right)} + p^{\left(bev\right)},~~f^{\left(ctr\right)} = f^{\left(ofs\right)} + f^{\left(bev\right)}
\end{equation}

\noindent where $p^{\left(bev\right)} \in \mathbb{R}^{HW \times 3}$ indicates the coordinates of points on BEV maps, the height is set to $0$ by default; $\mathcal{M}$ denotes a MLP layer; $\left[*;*\right]$ means the concatenation operation. 

To measure the quality of the shifted centers for sampling, we choose 3D centerness \cite{tian2019fcos,Yang20203dssd} as metric indicator, which can be written as,

\begin{equation}
s^{\left(ctrns\right)} = \sqrt[3]{\frac{\min\left(x_{f}, x_{b}\right)}{\max{\left(x_{f}, x_{b}\right)}} \times \frac{\min\left(y_{l}, y_{r}\right)}{\max{\left(y_{l}, y_{r}\right)}} \times \frac{\min\left(z_{t}, z_{b}\right)}{\max{\left(z_{t}, z_{b}\right)}} }
\end{equation}

\noindent where $\left(x_{f}, x_{b}, y_{l}, y_{r}, z_{t}, z_{b}\right)$ denotes the distance from candidate centers to front, back, left, right, top and bottom surfaces of the corresponding boxes they fall in. $s^{\left(ctrns\right)}$ is close to $1$ when the shifted candidate centers are more accurate. The centerness values for those outside the bounding boxes are $0$. During training and testing, we feed candidate center features $f^{\left(ctr\right)}$ into a MLP layer with a sigmoid function to predict this score, which is used as confidence score to sample high-quality centers with NMS, by treating each center as $1\times1\times1$ cube.

\subsection{Implicit Boundary Generation}
\label{subsec:generation}

After sampling candidate centers, we perform implicit functions on points in a local 3D space around each center. High-quality boundaries can be subsequently generated according to the assigned implicit values.

\vspace{0.05in}
\noindent \textbf{Virtual Sampling Strategy.}
Given a candidate center $p^{\left(ctr\right)}_{k}$, we get its surrounding local space by drawing a ball with radius $r$, and randomly select $m$ points from the space. The set of sampled points are defined as $\mathcal{B}^{p}_{k}=\mathcal{Q}\left(p^{\left(ctr\right)}_{k}\right)=
\left\{p_{i} \in \mathcal{P} ~\arrowvert~ \|p^{\left(ctr\right)}_{k} - p_{i} \|_{2} < r \right\}$, where $card\left(\mathcal{B}^{p}_{k}\right)=m$. We assign $r$ a relatively large value to ensure the ball covers as many points as possible. For sampled points in $\mathcal{B}^{p}_{k}$, we also gather their features from $f^{\left(point\right)}$ and denote them as $\mathcal{B}^{f_{p}}_{k}$.

\begin{figure}[tbp]
\centering
\includegraphics[scale=0.28]{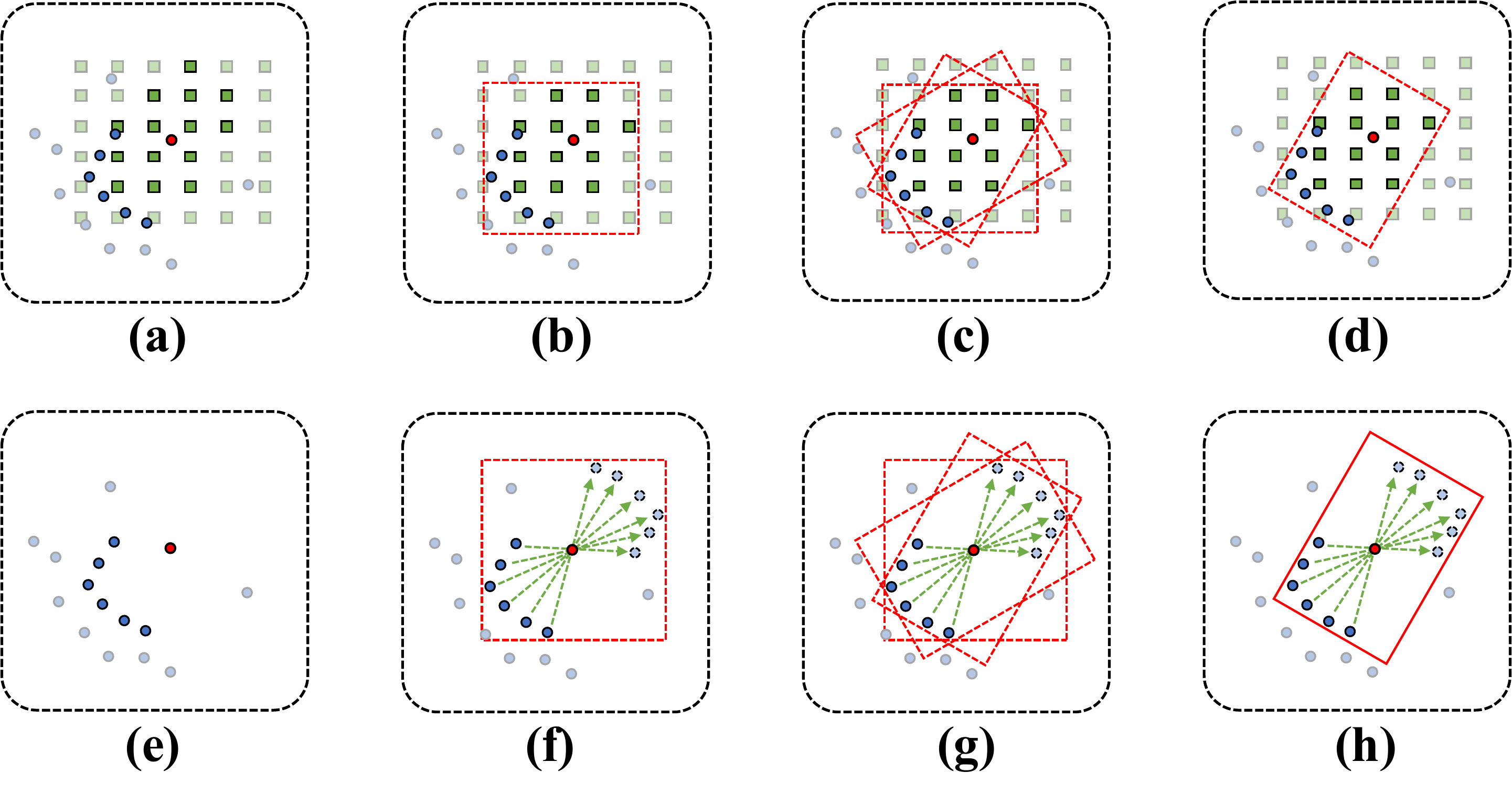}
\caption{The illustration of implicit boundary generation. Note that (a)-(d) represents the \emph{sampling} strategy, and (e)-(h) means the \emph{centrosymmetry} strategy. The red point denotes a candidate center, blue and green points are sampled raw points and virtual points. Particularly, darker color represents the inside points filtered by a threshold $t$. Red boxes are generated boundaries with different orientations. We omit virtual points in (e)-(h) for better view. \label{fig:minrec}}
\vspace{-0.1in}
\end{figure}

However, along with distance increase, point clouds become sparser and fewer points fall on the object's surface. For distant objects, the point coordinate information may be insufficient to predict boxes. To this end, we present a \emph{virtual sampling strategy} as shown in Fig.~\ref{fig:minrec}(a). Concretely, a set of virtual points $\mathcal{V}_{k}$ are uniformly placed around the candidate center $p^{\left(ctr\right)}_{k}$ with the grid size of $S\times S\times S$ and the interval of $\left(x_{s}, y_{s}, z_{s}\right)$. On account of less computation cost, we also randomly sample $m$ virtual points from $\mathcal{V}_{k}$. To get features of sampled virtual points, we apply K-Nearest Neighbor to interpolate virtual point features from voxel-wise features $f^{\left(v_{4}\right)}$ because of its large receptive field. A MLP layer is further employed to encode the interpolated features as well as their coordinates. Similarly, we denote the set of sampled virtual points and their features as $\mathcal{B}^{v}_{k}$ and $\mathcal{B}^{f_{v}}_{k}$.
Experiments in Tab.~\ref{tab:ablation_centerosymmetry} show that this simple yet effective strategy plays a key role in boundary generation, because it can not only fill the empty region, but also learn rich semantic information.

\vspace{0.05in}
\noindent \textbf{Implicit Function.}
Intuitively, whether a sampled point belongs to a box (\ie, inside the box) depends on its corresponding candidate center. The closer the euclidean or feature distances of two points are, the higher probability that they belong to the same box (object). Such a conditional relation inspires us to introduce an implicit function, which produces kernels conditioned on the candidate centers. The kernels further convolve with sampled points, so that the implicit values can be adjusted dynamically. More precisely, the generated kernels are reshaped as parameters of two convolution layers with the channel of $16$, the relative distance between the candidate center and sampled points are also involved. Take the sampled virtual points $\mathcal{B}^{v}_{k}$ as an example, the formulations are defined as,

\begin{equation}
\theta_{k} = \mathcal{M}\left(\left[f^{\left(ctr\right)}_{k};~p^{\left(ctr\right)}_{k} ~\right] \right)
\end{equation}
\vspace{-0.25in}

\begin{equation}
\mathcal{H}_{k}^{v} = sigmoid\left(\mathcal{O}\left(\left[\mathcal{B}^{f_{v}}_{k};~\mathcal{B}^{v}_{k} - p^{\left(ctr\right)}_{k} \right],~\theta_{k} \right) \right)
\end{equation}

\noindent where $\mathcal{H}_{k}^{v} \in \mathbb{R}^{1\times m}$ is the assigned implicit values; $\mathcal{O}\left(*;\theta\right)$ means the convolution operation with kernel $\theta$. The final implicit values $\mathcal{H}_{k}$ of candidate center $p^{\left(ctr\right)}_{k}$ is achieved by integrating outputs both from raw points $\mathcal{B}^{p}_{k}$ and virtual points $\mathcal{B}^{v}_{k}$.

\vspace{0.05in}
\noindent \textbf{Boundary Generation.}
By setting a threshold $t=0.5$, we can easily distinguish the inside and outside points with $\mathcal{H}$. The key challenge now is \emph{how to fit a boundary according to the classified points}. Generally, a regular boundary box in 3D space should include two factors: size and orientation.
	
For the size, we apply a strategy named \textit{`sampling'} to directly fit a minimum bounding box by using inside points, because (1) point clouds are mostly on the surface of objects; and (2) virtual points can significantly complement point clouds, reducing the sparsity in objects caused by distance or occlusion. Particularly, the center of the boundary can be easily computed, which may be different from the candidate center, as illustrated in Fig.~\ref{fig:minrec}(a)-(b). As a contrast, we also introduce an algorithm termed \emph{`centrosymmetry'} to first project the symmetric point of each inside point according to the candidate center\footnote{An object or its surface points are not centrosymmetric but the bounding box is.}, and then draw a minimum bounding box with both original and projected points, as shown in Fig.~\ref{fig:minrec}(e)-(f). Obviously, this strategy uses the parameter of center and the quality of the boundary depends on the accuracy of the candidate center. Experiments in Tab.~\ref{tab:ablation_centerosymmetry} clearly suggests that the boundary boxes generated by our proposed implicit fields are more robust.


For the orientation of objects in 3D object detection, it naturally ranges from $0$ to $2\pi$ and is usually not parallel to $x$-$y$ axes. Therefore, it is necessary to fit inside points better by rotating boundary boxes. Concretely, we first narrow down the search space from $\left[0, 2\pi\right)$ to $\left[0, \frac{\pi}{2}\right)$ (\ie, convert to the first quadrant) and then divide it into $h$ different angles, thereby producing $h$ different minimum bounding boxes with different angles. As a result, we accumulate the point-to-surface distance for each box and select the minimum one as the final boundary, shown in Fig.~\ref{fig:minrec}(c)-(d) and (g)-(h). We assign the rotation of the minimum one $r_{a} \in \left[0, \frac{\pi}{2}\right)$ as the boundary's orientation. Furthermore, denote the boundary size as $\left(l_{a}, w_{a}, h_{a}\right)$, we empirically correct the orientation and expand the range to $\left[0, \pi\right)$ by,

\begin{small}
\begin{equation}
r_{a} = \left\{\begin{aligned}
r_{a},&~~if~l_{a} \ge w_{a}
\\ 
r_{a} + \frac{\pi}{2},&~~otherwise
\end{aligned}\right.
\end{equation}
\end{small}

\subsection{Occupant Aggregation}
\label{subsec:aggregation}

As shown in Tab.~\ref{tab:ablation_recall}, boundary boxes predicted by our implicit boundary generation stage achieve the competitive recall performance. However, for 3D object detection, it still lacks the classification score and the accurate orientation (which should range from $\left[0, 2\pi\right)$). To this end, we reuse the implicit values $\mathcal{H}$ to refine the boundary boxes by aggregating features of inside points and suppressing the effect from outside points. Concretely, we uniformly sample $6 \times 6 \times 6$ grid points within each boundary box. Then, a set abstraction layer is applied to aggregate features of inside points as well as the voxel-wise features $f^{\left(v_{3}\right)}$ and $f^{\left(v_{4}\right)}$ at the location of each grid point. Finally, we concatenate all grid points' features and feed them into a detection head. The head is built with three branches for classification confidence, direction prediction and box refinement respectively. Particularly, each branch has four MLP layers with a channel of $256$ and shares the first two layers.

\subsection{Loss Function \label{subsec:loss}}

The overall loss functions are composed of six terms, \textit{i.e.}, the candidate shifting loss, the centerness confidence loss, the implicit function loss, the classification loss, the box refinement loss and the direction prediction loss,

\begin{equation}
\begin{aligned}
\mathcal{L} = \lambda_{1}\mathcal{L}_{ofs} + &\lambda_{2}\mathcal{L}_{ctrns} + \lambda_{3}\mathcal{L}_{imp} \\
 + &\lambda_{4}\mathcal{L}_{cls} + \lambda_{5}\mathcal{L}_{box} + \lambda_{6}\mathcal{L}_{dir}
\end{aligned}
\end{equation}

\noindent where $\lambda_{i}$ is the coefficient to balance each term. Similar to \cite{shi2020pv,deng2020voxel}, we empirically set $\lambda_{1} = \lambda_{2} = \lambda_{4} = 1.0$, $\lambda_{3} = \lambda_{5} = 2.0$ and $\lambda_{6} = 0.2$.

Here, we mainly describe the first three objectives proposed by us.
Denote the symbols with hat `$\wedge$' as ground truth, each formulation can be defined as,

\begin{small}
\begin{equation}
\mathcal{L}_{ofs} = \frac{1}{|\mathfrak{N}_{pixel}|} \sum_{i \in \mathfrak{N}_{pixel}} \mathcal{L}_{smooth_{L1}} \left(p^{\left(ofs\right)}_{i},~\widehat{p^{\left(ofs\right)}_{i}} \right)
\end{equation}
\end{small}
\vspace{-0.25in}

\begin{small}
\begin{equation}
\mathcal{L}_{ctrns} = \frac{1}{|\mathfrak{N}_{pixel}|} \sum_{i=1}^{HW} \mathcal{L}_{focal} \left(s^{\left(ctrns\right)}_{i},~\widehat{s^{\left(ctrns\right)}_{i}} \right)
\end{equation}
\end{small}
\vspace{-0.25in}

\begin{small}
\begin{equation}
\mathcal{L}_{imp} = \frac{1}{|\mathfrak{N}_{center}|} \sum_{i \in \mathfrak{N}_{center}} \mathcal{L}_{BCE} \left(\mathcal{H}_{i},~\widehat{\mathcal{H}_{i}} \right)
\end{equation}
\end{small}

\noindent where $\mathfrak{N}_{pixel}$ and $\mathfrak{N}_{center}$ indicate the set of indices of positive pixels/candidate centers if they are inside objects' bounding boxes; `$|\cdot|$' means the cardinality.

\begin{figure*}[thb]
\begin{centering}
\includegraphics[width=0.9\linewidth]{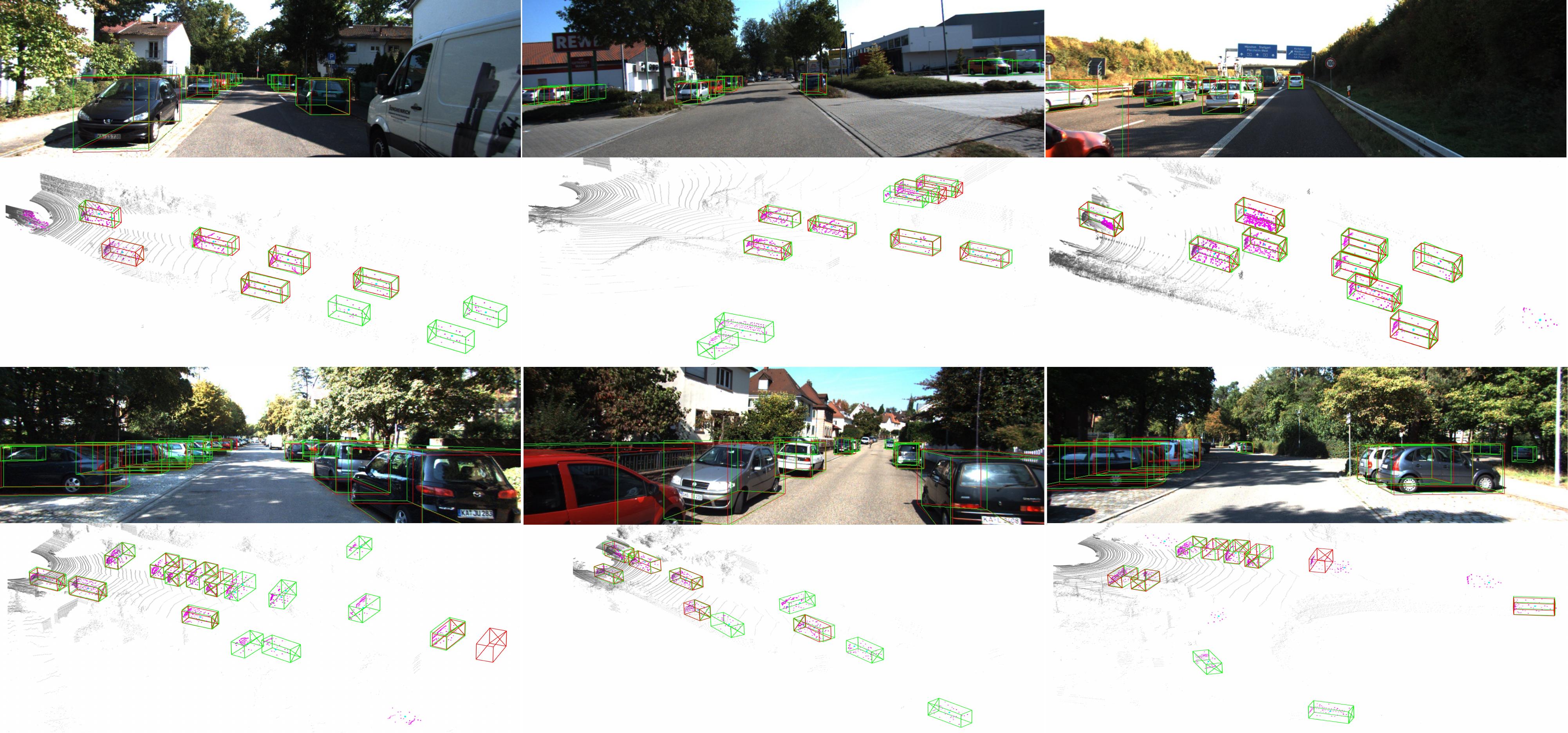}
\caption{Visualization on KITTI \emph{val} set. The ground truth boxes and our predicted bboxes are drew in red and green. The internal raw points and virtual points predicted by implicit functions are highlighted in purple.}
\label{fig:vis_results}
\end{centering}
 \vspace{-0.8em}
\end{figure*}

\section{Experiments}
\subsection{Dataset and Protocols}
To verify the efficacy of our proposed model, we evaluate it on two popular public benchmarks, KITTI 3D detection benchmark~\cite{geiger2013vision} and Waymo Open Dataset~\cite{sun2020scalability} (WOD).

\vspace{0.05in}
\noindent \textbf{KITTI Setup.}
The KITTI dataset contains $7,481$ training frames and $7,518$ testing frames in autonomous driving scenes. Following the standard setting, the training data are divided into a \textit{train} set with $3,712$ samples and a \textit{val} set with $3,769$ samples. We report the mean average precision of 3D object detection ($\rm AP_{3D}$) and bird’s eye view ($\rm AP_{BEV}$) on both the \textit{val} set and online \textit{test} server. 
For fair comparison, the 40 recall positions based metric $\rm AP|_{R40}$ is reported on \textit{test} server while $\rm AP|_{R11}$ with 11 recall positions is reported on \textit{val} set.
On the KITTI benchmark, according to the object size, occlusion ratio, and truncation level, the task can be categorized into `Easy', `Mod.' and `Hard', we report the results in all three tasks, and ranks all methods based on the $\rm AP_{3D}$ of `Mod.' setting as in KITTI benchmark.
In particular, we focus on the `Car' category as many recent works \cite{deng2020voxel,zhou2020joint,miao2021pvgnet} and adopt IoU = $0.7$ for evaluation.
When performing experimental studies on the \emph{val} set, we use the train data for training. For the \textit{test} server, we randomly select $80\%$ samples for training and use the remaining $20\%$ data for validation. 

\vspace{0.05in}
\noindent \textbf{Waymo Setup.}
We also conduct experiments on the recently released large-scale diverse dataset, Waymo Open Dataset ~\cite{sun2020scalability}, to verify the generalization of our method. The dataset collects RGB images and 3D point clouds from five high-resolution cameras and LiDAR sensors, respectively.
It provides annotated 798 training sequences, 202 validation sequences from different scenes, and another 150 test sequences without labels.
For evaluation, we adopt the officially released evaluation to calculate the average precision (AP) and average precision weighted by heading (APH).
Specifically, two levels are set according to different LiDAR points included by objects. And three distance (0 - 30m, 30 - 50m, 50m - $\infty$) to sensor are considered under each level.

\subsection{Implementation Details}
\vspace{0.05in}
\noindent \textbf{Network Structure.}
On KITTI dataset, the detection range is limited to $\left(0, 70.4\right)m$ for the $x$ axis, $\left(-40, 40\right)m$ for the $y$ axis, and $\left(-3, 1\right)m$ for the $z$ axis. Before taken as input of our ImpDet, the raw point clouds are divided into regular voxels with voxel size of $\left(0.05, 0.05, 0.1 \right)m$. 
As for Waymo Open Dataset, the range of point clouds is clipped into $\left(-75.2, 75.2\right)m$ for both the $x$ and $y$ axes, and $\left(-2, 4\right)m$ for the $z$ axis. The voxel size is $\left(0.1, 0.1, 0.15 \right)m$.
For these two datasets, each voxel randomly samples at most $5$ points. We stacked two VFE layers with filter numbers of $32$ and $64$ to extract point-wise features. In regard to the backbone network, there are totally five 3D sparse convolution blocks with the output channels of $32$, $32$, $64$, $64$, and $128$, respectively. 
Following \cite{wang2019voxel}, we adopt 2 convolutional layers and 2 deconvolutional layers as FPN structure. The output feature dimension is $128$ and $256$ for KITTI and Waymo Open Dataset, respectively.

\vspace{0.05in}
\noindent \textbf{Hyper Parameters.}
After the candidate shifting layer, we select top-$512$ candidate centers for the following stage. The number of sampled points for implicit fields is set to $m=256$ with the radius $r=3.2m$. For virtual sampling strategy, we empirically assign $10\times 10\times 10$ as grid size, the interval is $\left(0.6, 0.6, 0.3\right)m$. During implicit boundary generation, we choose the optimal boundary by enumerating $h=7$ angles from $\left[0,~\frac{\pi}{2} \right)$. All of these settings are applied to both datasets.

\vspace{0.05in}
\noindent \textbf{Training.}
Our framework is built on OpenPCDet codebase \cite{od2020openpcdet}. We train the whole model with batch size as $3$ and learning rate as $0.01$ on $8$ Tesla V100 GPUs. Adam optimizer is adopted to train our model for totally $80$ and $60$ epochs on KITTI and Waymo Open Datasets, respectively. Widely-used data augmentation strategies like flipping, rotation, scaling, translation and sampling are also adopted. 

\vspace{0.05in}
\noindent \textbf{Inference.}
During inference, we first filter the predicted boxes with $0.3$ confidence threshold and then perform NMS with $0.1$ IoU threshold to remove the redundant predictions. 


\begin{table}[tbp]
\begin{centering}
\small{
\setlength{\tabcolsep}{1.7mm}{
\begin{tabular}{l|c|ccc}
\hline 
\multicolumn{1}{c|}{\multirow{2}{*}{Method}} &
\multicolumn{1}{c|}{\multirow{2}{*}{Reference}} &\multicolumn{3}{c}{$\rm AP_{3D}$} \tabularnewline
 && \textbf{Mod.} & Easy & Hard \tabularnewline
\hline
\hline
VoxelNet \cite{zhou2018voxelnet} & CVPR 2018 & 64.17 & 77.82 & 57.51 \tabularnewline
PointPillars \cite{lang2019pointpillars} & CVPR 2019 & 74.31 & 82.58 & 68.99 \tabularnewline
SECOND \cite{yan2018second} & Sensors 2018 & 75.96 & 84.65 & 68.71 \tabularnewline
Patches \cite{lehner2019patch} & NeurIPS 2019 & 77.20 & 88.67 & 71.82 \tabularnewline
HVPR \cite{noh2021hvpr}& CVPR 2021  & 77.92 & 86.38 & 73.04 \tabularnewline
3DSSD \cite{Yang20203dssd} & CVPR 2020 & 79.57 & 88.36 & 74.55 \tabularnewline

RangeIoUDet \cite{liang2021rangeioudet}& CVPR 2021  & 79.80 & 88.60 & 76.76 \tabularnewline
CIA-SSD \cite{zheng2020cia} & AAAI 2021 & 80.28 & 89.59 & 72.87 \tabularnewline
Voxel R-CNN \cite{deng2020voxel} & AAAI 2021 & 81.62 & \textbf{90.90} & 77.06 \tabularnewline
\hline
PI-RCNN \cite{xie2020pi} & AAAI 2020 & 74.82 & 84.37 & 70.03 \\ 
PointRCNN \cite{shi2019pointrcnn} & CVPR 2019 & 75.64 & 86.96 & 70.70 \tabularnewline
MMLab-Part$\rm A^2$ \cite{shi2019part} & Arxiv 2019 & 78.49 & 87.81 & 73.51 \tabularnewline 
SERCNN \cite{zhou2020joint} & CVPR 2020 & 78.96 & 87.74 & 74.30 \tabularnewline
STD \cite{yang2019std} & ICCV 2019 & 79.71 & 87.95 & 75.09 \tabularnewline
SA-SSD \cite{he2020structure} & CVPR 2020 & 79.79 & 88.75 & 74.16 \tabularnewline
PV-RCNN \cite{shi2020pv} & CVPR 2020 & 81.43 & 90.25 & 76.82 \tabularnewline

\hline
ImpDet(Ours)& -  & \textbf{82.14} & 88.39 & 76.98 \tabularnewline
\hline
\end{tabular}}}
\par\end{centering}
\caption{Comparison with the state-of-the-art competitors on KITTI \emph{test} split. Methods are grouped into two categories: without (top) or with (bottom) segmentation branch.
\label{Tab: Test_Results}}
\vspace{-1.5em}
\end{table}

\begin{table*}[tbp]
\begin{centering}
\setlength{\tabcolsep}{5mm}{
\begin{tabular}{c|ccccc}
\hline
 \multirow{2}{*}{Method} &   \multicolumn{4}{c}{$\rm LEVEL\_1(AP/APH)$}\\
 & \textbf{Overall}  & 0 - 30m          & 30 - 50m          & 50 - $ \infty$  \\
\hline
\hline
  
  PointPillars \cite{lang2019pointpillars} &  56.62 /\quad-\quad\quad & 81.01 /\quad-\quad\quad & 51.75 /\quad-\quad\quad & 27.94 /\quad-\quad\quad \\
  MVF \cite{zhou2020end} & 62.93 /\quad-\quad\quad & 86.30 /\quad-\quad\quad & 60.02 /\quad-\quad\quad & 36.02 /\quad-\quad\quad \\
 

PV-RCNN \cite{shi2020pv} & 70.30 / 69.69 & 91.92 / 91.34 & 69.21 / 68.53 & 42.17 / 41.31 \\


PVGNet \cite{miao2021pvgnet} & 74.00 /\quad-\quad\quad & \quad-\quad /\quad-\quad\quad & \quad-\quad /\quad-\quad\quad & \quad-\quad /\quad-\quad\quad \\


\hline
ImpDet(Ours)&  \textbf{74.38 / 73.87} & \textbf{91.98 / 91.52}
 & \textbf{72.86 / 72.29} & \textbf{49.13 / 48.45}\\
\hline
\end{tabular}}
\par\end{centering}
\caption{Performance comparison on WOD \emph{val} split. We report all distance ranges results on vehicle category.
\label{Tab: Val_Results_Waymo}}
\vspace{-0.5em}
\end{table*}

\begin{table}[tbp]
\begin{centering}
\small{
\setlength{\tabcolsep}{0.8mm}{

\begin{tabular}{c|ccc}
\hline
 \multirow{2}{*}{Method} &  \multicolumn{3}{c}{$\rm AP_{3D} / AP_{BEV} $} \\
&   \textbf{Mod.}  & Easy  & Hard  \\
\hline
\hline

 VoxelNet \cite{zhou2018voxelnet} &  65.46 / 84.81 & 81.97 / 89.60 & 62.85 / 78.57\\
SECOND \cite{yan2018second} &  76.48 / 87.07 & 87.43 / 89.96 & 69.10 / 79.66\\
 PointPillars \cite{lang2019pointpillars} &  76.99 / 87.06 & 87.29 / 90.07 & 70.84 / 83.81\\
  Patches \cite{lehner2019patch} &  79.04 /\quad-\quad\quad & 89.55 /\quad-\quad\quad & 78.10 /\quad-\quad\quad\\
 3DSSD \cite{Yang20203dssd} &  79.45 /\quad-\quad\quad & 89.71 /\quad-\quad\quad & 78.67 /\quad-\quad\quad\\
 
 CIA-SSD \cite{zheng2020cia} & 79.81 /\quad-\quad\quad & 90.04 /\quad-\quad\quad & 78.80 /\quad-\quad\quad\\

 RangeIoUDet \cite{liang2021rangeioudet} & 81.36 /\quad-\quad\quad & 89.32 /\quad-\quad\quad & 78.29 /\quad-\quad\quad\\
 HVPR \cite{noh2021hvpr} &  82.05  /\quad-\quad\quad & \textbf{91.14} /\quad-\quad\quad & \textbf{79.49} /\quad-\quad\quad\\

 Voxel R-CNN \cite{deng2020voxel} & 84.52 /\quad-\quad\quad & 89.41 /\quad-\quad\quad & 78.93 /\quad-\quad\quad\\
\hline
 PI-RCNN \cite{xie2020pi} & 78.53 /\quad-\quad\quad & 88.27 /\quad-\quad\quad & 77.75 /\quad-\quad\quad \\ 
 PointRCNN \cite{shi2019pointrcnn} &  78.63 / 87.89 & 88.88 / 90.21 & 77.38 / 85.51\\
 SERCNN \cite{zhou2020joint} & 79.21 / 87.53 & 89.50 / 90.23 & 78.16 / 86.45 \\
 MMLab-Part$\rm A^2$ \cite{shi2019part} &  79.47 / 88.61 & 89.47 / 90.42 & 78.54 / 87.31\\ 
 STD \cite{yang2019std} &  79.80 / 88.50 & 89.70 / \textbf{90.50} & 79.30 / 88.10\\
  SA-SSD \cite{he2020structure} &  79.99 /\quad-\quad\quad & 90.15 /\quad-\quad\quad & 78.78 /\quad-\quad\quad\\
 P2V-RCNN \cite{li2021p2v} &  82.49 /\quad-\quad\quad   & 86.83 /\quad-\quad\quad & 77.61 /\quad-\quad\quad \\
 PV-RCNN \cite{shi2020pv} &  83.90 /\quad-\quad\quad  & \quad-\quad/\quad-\quad\quad  & \quad-\quad/\quad-\quad\quad  \\
 
\hline
ImpDet(Ours)&  \textbf{85.38} / \textbf{89.03} & 89.91 / \textbf{90.50} & 79.25 / \textbf{88.24}\\
\hline
\end{tabular}}}
\par\end{centering}
\caption{Performance comparison on KITTI \emph{val} split.
Methods are grouped into two categories: without (top) or with (bottom) segmentation branch.
\label{Tab: Val_Results}}
\vspace{-0.2in}
\end{table}

\begin{table}[tbp]
\begin{centering}
\small{
\setlength{\tabcolsep}{1.6mm}{
\begin{tabular}{l|ccc}
\hline
\multicolumn{1}{c|}{$\rm AP_{3D}$} & Mod. & Easy & Hard \tabularnewline
\hline
PV-RCNN \cite{shi2020pv} & 70.47 / 57.90  & - / - & - / - \tabularnewline
Ours & 72.38 / 64.63 & 89.25 / 69.58 & 69.59 / 59.14 \tabularnewline
\hline
\end{tabular}}}
\par\end{centering}
\caption{Performance comparison of Cyclist / Pedestrian categories on KITTI \emph{val} set with R11. \label{tab:ablation_person}}
\end{table}

\begin{table}[tbp]
\begin{centering}
\small{
\setlength{\tabcolsep}{1mm}{
\begin{tabular}{l|ccc}
\hline
\multicolumn{1}{c|}{Method} & PV-RCNN$^{*}$ \cite{shi2020pv} & Voxel R-CNN$^{*}$ \cite{deng2020voxel} & IBG \tabularnewline
\hline
\multicolumn{1}{c|}{Recall (IoU=0.7)} & 76.40 & 77.10 & \textbf{77.78} \tabularnewline
\hline
\end{tabular}}}
\par\end{centering}
\caption{Comparison of recall using different proposal generation networks. `IBG' denotes our implicit boundary generation and * indicates our reproduced performance. \label{tab:ablation_recall}}
\end{table}

\subsection{Comparison with State-of-the-Arts}

\vspace{0.05in}
\noindent \textbf{KITTI \emph{test} Split.}
To verify the efficacy of our ImpDet, we evaluate our model on KITTI online \textit{test} server. As shown in Tab. \ref{Tab: Test_Results}, we report the $\rm AP_{3D}$ results over three settings. From the table, we can observe that: 
(1) It is obvious that our model can achieve state-of-the-art performance compared with previous methods on the most concerned `Mod.' setting. This demonstrates the efficacy of our motivation, which leverages the implicit fields to fit high-quality and robust boundaries without any pre-defined anchors for 3D object detection.
(2) We group existing methods in tables based on whether containing a segmentation branch. As can be seen, the performance improvement of our ImpDet over the existing 3D object detectors with segmentation branch is significant. Concretely, we achieve $0.33\%/1.97\%$ higher accuracy on `Mod.' setting than PV-RCNN \cite{shi2020pv} and SA-SSD \cite{he2020structure}. It proves that our implicit field learning has the potential capacity in applying to 3D object detection task. 
(3) We observe that our model get inferior results on easy cases. One possible reason is that there is a trade-off between memory footprint and accuracy during sampling, which is harsh for easy cases (with thousands of foreground points).


\vspace{0.05in}
\noindent \textbf{KITTI \emph{val} Split.}
We also compare our method with competitors over the KITTI \textit{val} set. As shown in Tab. \ref{Tab: Val_Results}, our ImpDet can achieve state-of-art performance. Especially, ImpDet outperforms the previous best significantly, \textit{e.g.}, $0.86\%$ over Voxel R-CNN \cite{deng2020voxel} and $1.48\%$ over PV-RCNN \cite{shi2020pv} on `Mod.' setting.
Similar conclusions are drawn in Tab.~\ref{tab:ablation_person}, which lists the results of other categories, such as pedestrian and cyclist. It suggests that our sampling strategy also works well for small categories.
We also show some prediction results in Fig. \ref{fig:vis_results} and we project the 3D bounding boxes detected from LiDAR to the RGB images for better visualization. As observed, our ImpDet can produce high-quality 3D bounding boxes via implicit functions in different kinds of scenes. Remarkably, when there are fewer points on objects, our proposed virtual sampling strategy can significantly fill the empty region and thus assist in boundary generation with the assigned implicit values. 
Our ImpDet may fail on some cases if a candidate center is generated over a large empty area. The sampled virtual points cannot learn enough semantic features from their neighbor raw points.

\vspace{0.05in}
\noindent \textbf{Waymo \emph{val} Split.}
Table~\ref{Tab: Val_Results_Waymo} reports the vehicle detection results with 3D AP/APH on validation sequences. Without bells and whistles, our proposed method outperforms all existing state-of-the-art methods on the vehicle category. Improvements on all distance ranges indicate that our methods can robustly represent 3D object bounding boxes containing a various density of points. Especially, a larger gain has been achieved compared with PV-RCNN \cite{shi2020pv} on distance (50m - $\infty$), which illustrates that our implicit field learning performs better than directly parameters learning of bounding box with sparse points.

\begin{table}[tbp]
\begin{centering}
\small{
\setlength{\tabcolsep}{0.7mm}{
\begin{tabular}{l|ccc}
\hline
\multicolumn{1}{c|}{Method} & PA (1/0) & IoU (1/0) & $\rm AP_{3D}$ (Mod./Easy/Hard) \tabularnewline
\hline
\multicolumn{1}{c|}{\textit{w/o} cond.} & 71.38 / 95.91 & 57.46 / 91.49 & 76.21 / 85.98 / 68.19 \tabularnewline
\multicolumn{1}{c|}{\textit{w/o} dist.} & 88.36 / 97.04 & 75.38 / 95.13 & 84.82 / 89.46 / 78.79 \tabularnewline
\multicolumn{1}{c|}{Ours} & 89.82 / 97.20 & 77.28 / 95.53 &  85.38 / 89.91 / 79.25 \tabularnewline
\hline
\end{tabular}}}
\par\end{centering}
\caption{Ablation study of the design in implicit function. `\textit{w/o} dist.' denotes the relative distance is not involved.
`\textit{w/o} cond.' denotes the vanilla convolution layers with sampled point features and relative distances as inputs.
\label{tab:ablation_seg}}
\end{table}

\subsection{Ablation Study}

We conduct extensive ablation experiments to explore the effectiveness of different components in our ImpDet and analyze the contributions of implicit fields in 3D object detection. Models are trained on KITTI \textit{train} split and evaluated on the corresponding \textit{val} split. The results of car on the moderate task are reported with R11.

\vspace{0.05in}
\noindent \textbf{Analysis on Implicit Function.} To validate the effectiveness of our design in the implicit function, we
conduct several variants and apply both detection metric (AP$\rm_{3D}$) and segmentation metrics (Pixel Accuracy and IoU). For PA and IoU, we report both results on the categories of $0$ and $1$. 
Tab.~\ref{tab:ablation_seg} shows that (1) When the relative distance is not involved in the convolution layer (termed `\textit{w/o} dist.'), the performance drops a lot; (2) By directly using the vanilla convolution layers with sampled point features and relative distances as input (termed `\textit{w/o} cond.'), it gets much worse results. Those suggest the superiority of our design in the implicit function, and the better accuracy of implicit values facilitates much higher performance of object detection.


\vspace{0.05in}
\noindent \textbf{Analysis of Boundary Generation.} In Tab.~\ref{tab:ablation_centerosymmetry}, we first compare the performance with two boundary generation strategies, \textit{i.e.}, \textit{sampling} and \textit{centrosymmetry}. `point+virtual' means we utilize both sampled raw points and virtual points for boundary generation. First of all, we observe that additionally using virtual points can boost the performance by a large margin of $4.9\%$ and $14.73\%$ on both strategies. It clearly demonstrates the effectiveness of our proposed virtual sampling strategy in boundary generation, which can significantly fill empty regions in objects. Second, the \textit{sampling} strategy only with raw points achieves the worst results of $70.65/79.27\%$ on AP$\rm_{3D/BEV}$, we explain that too sparse point clouds may make the implicit fields inapplicable since there is no enough points to fit a boundary. Third, our \textit{sampling} strategy outperforms the \textit{centrosymmetry} by $1.05\%$ and $0.55\%$ on 3D and BEV accuracy. Recall the difference between these two strategies, the \textit{centrosymmetry} strategy additional needs the predicted center to perform the centrosymmetric projection for each point, thereby it strongly shows the robustness of our proposed implicit fields, even with some outliers. 

We also discuss the values of $h$ in Tab.~\ref{tab:num_angle}. As expected, if the division of angles is too large, it cannot fit a boundary well, resulting in a drop of detection performance. On the contrary, the more angles we divide, the less the accuracy gains and the higher the computation costs. We choose the optimal value when the model achieves the best performance, \ie, $h=7$.

Finally, to validate the quality of the predicted boundary boxes via implicit fields, we compute the recall rate with the ground-truth boxes.
To be fair, we show the recall rates for all methods with top-$100$ proposals on the car category over all difficulty levels.
As shown in Tab.~\ref{tab:ablation_recall}, only with the supervision of center coordinates, our introduced implicit field achieves a competitive result with $77.78\%$ recall rate, outperforming both PV-RCNN \cite{shi2020pv} and Voxel R-CNN \cite{deng2020voxel}. This indicates that our implicit field learning can robustly fit high-quality bounding boxes.

\begin{table}[tbp]
\begin{centering}
\small{
\setlength{\tabcolsep}{1.5mm}{
\begin{tabular}{l|cccc}
\hline
\multicolumn{1}{c|}{\multirow{2}{*}{Method}} & \multicolumn{2}{c|}{\emph{centrosymmetry}}  & \multicolumn{2}{c}{\emph{sampling}} \tabularnewline
\cline{2-5}
& point & \multicolumn{1}{c|}{point + virtual} & point & point + virtual \tabularnewline
\hline
AP$_{3D}\left(\%\right)$ & 79.43 & \multicolumn{1}{c|}{84.33} & 70.65 & 85.38 \tabularnewline
AP$_{BEV}\left(\%\right)$ & 87.81 & \multicolumn{1}{c|}{88.48} & 79.27 & 89.03 \tabularnewline
\hline
\end{tabular}}}
\par\end{centering}
\caption{Performance comparisons of different boundary generation strategies with both AP$\rm_{3D}$ and AP$\rm_{BEV}$.
\label{tab:ablation_centerosymmetry}}
\end{table}

\begin{table}[tbp]
\begin{centering}
\small{
\setlength{\tabcolsep}{4.1mm}{
\begin{tabular}{l|cccc}
\hline
\multicolumn{1}{c|}{$h$} & 3 & 5 & 7 & 9 \tabularnewline
\hline
AP$\rm_{3D}\left(\%\right)$ & 84.12 & 85.22 & 85.38 & 85.28 \tabularnewline
\hline
\end{tabular}}}
\par\end{centering}
\caption{Performance comparisons with different number of angle partition on AP$\rm_{3D}$. The best performance is achieved when $h=7$.
\label{tab:num_angle}}
\vspace{-0.2in}
\end{table}

\begin{table}[tbp]
\begin{centering}
\small{
\setlength{\tabcolsep}{1.9mm}{
\begin{tabular}{cccccc|c}
\hline
$f^{\left(v_{3}\right)}$  & $f^{\left(v_{4}\right)}$ & $\mathcal{B}^{f_{p}}$ & $\mathcal{B}^{f_{v}}$ & $\mathcal{H}^{p}\mathcal{B}^{f_{p}}$ & $\mathcal{H}^{v}\mathcal{B}^{f_{v}}$ & AP$\rm_{3D}\left(\%\right)$ \tabularnewline
\hline
\checkmark & \checkmark & & & & & 84.29 \tabularnewline
\checkmark & \checkmark & \checkmark & & & & 85.05 \tabularnewline
\checkmark & \checkmark &  & \checkmark & & & 84.85
\tabularnewline
\checkmark & \checkmark & & & \checkmark & & 85.10 \tabularnewline
\checkmark & \checkmark & & & & \checkmark &  85.16 \tabularnewline
\checkmark & \checkmark & & & \checkmark & \checkmark & \textbf{85.38} \tabularnewline
\hline
\end{tabular}}}
\par\end{centering}
\caption{Ablation study of different feature choices in occupant aggregation. \label{tab:ablation_occupant}}
\vspace{-1.em}
\end{table}

\vspace{0.05in}
\noindent \textbf{Analysis of Occupant Aggregation.}
In order to explore the contribution of our occupant aggregation module, we do experiments with different combinations of voxel-wise features ($f^{\left(v_{3}\right)}$ and $f^{\left(v_{4}\right)}$), sampled point features ($\mathcal{B}^{f_{p}}$), sampled virtual point features ($\mathcal{B}^{f_{v}}$) and those with implicit values ($\mathcal{H}^{p}\mathcal{B}^{f_{p}}$ and $\mathcal{H}^{v}\mathcal{B}^{f_{v}}$). As shown in Tab.~\ref{tab:ablation_occupant}, 
the comparisons between $\mathcal{B}^{f_{p}}$ and $\mathcal{H}^{p}\mathcal{B}^{f_{p}}$ or $\mathcal{B}^{f_{v}}$ and $\mathcal{H}^{v}\mathcal{B}^{f_{v}}$ consistently proves that the implicit values can effectively enhance the features of inside points, suggesting a solid advantage of incorporating implicit fields into 3D object detection. Interestingly, we observe that the virtual point features contribute more to the performance when $\mathcal{H}^{v}$ is applied (the second/third row and the fourth/fifth row). One possible explanation is that virtual points contain both rich semantic features and confused geometric features since they are randomly sampled in the 3D space. With the cooperation of implicit values, we can successfully suppress the distracting information. Moreover, the result from the last row demonstrates the complementarity of raw points and virtual points.

\section{Conclusion and Discussion}

In this paper, we introduce a new perspective to represent 3D bounding boxes with implicit fields. Our proposed framework, dubbed Implicit Detection or ImpDet, leverages the implicit function to generate high-quality boundaries by classifying points into two categories, \ie, inside or outside the boundary. A virtual sampling strategy is consequently designed to fill the empty regions around objects, making the boundary generation more robust. Our approach achieves comparable results to the current state-of-the-art methods both on KITTI and WOD benchmarks. 

ImpDet also encounters some challenges, including the trade-off between the computation cost and accuracy when sampling points in the local 3D space, and the results on easy objects. Nevertheless, we believe that this work can be inspiring and helpful for encouraging more researches.


{\small
\bibliographystyle{cvpr22}
\bibliography{cvpr22}
}

\end{document}